\documentclass[twocolumn,journal]{IEEEtran}

\usepackage{comment}
\usepackage{tikz}
\usepackage{graphicx}
\usepackage{amsmath}
\usepackage{amsfonts}
\usepackage{amssymb}
\usepackage{mathrsfs}
\usepackage{fix2col}
\usepackage{latexsym}
\usepackage[dvips]{epsfig}
\usepackage[font=footnotesize]{subfig}
\usepackage{hyperref}
\usepackage{cite}
\usepackage{graphics}
\usepackage{epstopdf}
\usepackage{enumerate}
\usepackage{cite}
\usepackage{epstopdf}
\usepackage{arydshln}
\usepackage{multirow}

\newtheorem{theorem}{Theorem}
\newtheorem{definition}{Definition}
\newtheorem{lemma}{Lemma}
\newtheorem{remark}{Remark}
\newtheorem{corollary}{Corollary}

\newtheorem{example}{Example}
\newtheorem{algor}{Algorithm}
\newtheorem{proof}{Proof}

\begin {document}
\global\long\def\S{\mathbb{S}}
\global\long\def\R{\mathbb{R}}
\global\long\def\Edges{\mathcal{E}}
\global\long\def\Tree{\mathcal{T}}

\title{  Sparse Inverse Covariance Estimation for Chordal Structures
\thanks{Email: fattahi@berkeley.edu, ryz@berkeley.edu, and sojoudi@berkeley.edu.}}
\author{
	Salar Fattahi, Richard Y. Zhang, and Somayeh Sojoudi
\thanks{Salar Fattahi and Richard Y. Zhang are with the Department of Industrial Engineering and Operations Research, University of California, Berkeley. Somayeh Sojoudi is with the Departments of Electrical Engineering and Computer Sciences and Mechanical Engineering, University of California, Berkeley. This work was supported by the ONR grant N00014-17-1-2933, DARPA grant D16AP00002, and AFOSR grant FA9550-17-1-0163.}}
 \maketitle
 
 \begin{abstract}
 	In this paper, we consider the Graphical Lasso (GL), a popular optimization problem for learning the sparse representations of high-dimensional datasets, which is well-known to be computationally expensive for large-scale problems. Recently, we have shown that the sparsity pattern of the optimal solution of GL is equivalent to the one obtained from simply thresholding the sample covariance matrix, for sparse graphs under different conditions. We have also derived a closed-form solution that is optimal when the thresholded sample covariance matrix has an acyclic structure. As a major generalization of the previous result, in this paper we derive a closed-form solution for the GL for graphs with chordal structures. We show that the GL and thresholding equivalence conditions can significantly be simplified and are expected to hold for high-dimensional problems if the thresholded sample covariance matrix has a chordal structure. We then show that the GL and thresholding equivalence is enough to reduce the GL to a maximum determinant matrix completion problem and drive a recursive closed-form solution for the GL when the thresholded sample covariance matrix has a chordal structure. For large-scale problems with up to 450 million variables, the proposed method can solve the GL problem in less than 2 minutes, while the state-of-the-art methods converge in more than 2 hours. 
 \end{abstract}

\section{Introduction}

In recent years, there has been a great deal of interest in developing computationally efficient methods to analyze large-scale and high-dimensional data. The data collected in practice is often overwhelmingly large. Therefore, designing simple, yet informative models for describing the underlying structure of data is of significant importance. Hence, sparsity-promoting techniques have become an essential part of inference and learning methods. These techniques have been widely-used in data mining \cite{Garcke01}, pattern recognition \cite{Wright10}, functional connectivity of the human brain \cite{Sojoudi14}, distributed controller design \cite{Fardad11, SODC2016}, transportation systems~\cite{Salar17}, and compressive sensing \cite{Candes07}. On the other hand, in many applications, the number of available data samples is much smaller than the dimension of the data. This implies that most of the statistical learning techniques, which are proven to be consistent with the true structure of the data fail dramatically in practice. This is due to the fact that most of the convergence results in the inference methods are contingent upon the availability of a sufficient number of samples, which may not be the case in practice. In an effort to overcome this issue, sparsity-inducing penalty functions are often used to arrive at a parsimonious graphical model for the available data. Graphical Lasso (GL) \cite{Friedman08, Banerjee08} is one of the most widely-used methods for sparse estimation of the inverse covariance matrices via the augmentation of a Lasso-type penalty function. It is known that the GL can be computationally prohibitive for the large-scale problems, which limits its applicability in practice. Recently, it has been shown in various applications, such as brain connectivity networks, electrical circuits, and transportation networks, that the thresholding technique and the GL lead to the same sparsity structure~\cite{Sojoudi16, Salar17}. Moreover, \cite{Sojoudi16} shows that under some conditions, a simple thresholding of the sample covariance matrix will result in the same sparsity pattern as the optimal solution of the GL. These conditions have been modified in~\cite{Salar17} to depend only on the sample covariance matrix (and not the optimal solution of the GL). Based on this equivalence,~\cite{Salar17} introduces a closed-form solution for the GL, the exactness of which depends on the sparsity structure of the thresholded sample covariance matrix. In another line of work, \cite{Mazumdar12} and \cite{Witten11} consider the disjoint components of the thresholded sample covariance matrix and show that the GL can be solved independently for each of the disjoint components. Although this result does not require additional conditions on the structure, its applicability is limited since it does not reveal any information about the connectivity of the sparsity graph corresponding to the optimal solution of the GL.
\subsection{Problem Formulation}

Consider a random vector $\mathcal{X} = [x_1,x_2,...,x_d]$ with an underlying multivariate Gaussian distribution. Let $\Sigma_*$ denote the covariance matrix of this random vector. Without loss of generality, we assume that $\mathcal{X}$ has a zero mean. The goal is to estimate the entries of $\Sigma_*^{-1}$ based on $n$ independent samples $\mathbf{x}_{(1)},\mathbf{x}_{(2)},...,\mathbf{x}_{(n)}$ of $\mathcal{X}$. The sparsity pattern of $\Sigma_*^{-1}$ determines which random variables  in $\mathcal{X}$ are  conditionally independent. In particular, if the $(i,j)^{\text{th}}$ entry of $\Sigma_*^{-1}$ is zero, it means that $x_i$ and $x_j$ are conditionally independent, given the remaining entries of $\mathcal{X}$ (the value of this entry is proportional to the partial correlation between $x_i$ and $x_j$). In this paper, we assume that $\Sigma_*^{-1}$ is sparse and non-singular. The problem of studying  the conditional independence of different entries of $\mathcal{X}$ is hard in practice due to the fact that the true covariance matrix is rarely known \textit{a priori}. Therefore, the sample covariance matrix must instead be used to estimate the true covariance matrix. Let $\Sigma$ denote the sample covariance matrix. To estimate $\Sigma_*^{-1}$, consider the  optimization problem
\begin{equation}\label{opt_inv}
\underset{
	\begin{subarray}{c}
	S\in\mathbb{S}^d_+
	\end{subarray}
}{\text{minimize}} \ -\log\det(S)+\text{trace}(\Sigma S)
\end{equation}
The optimal solution of the above  problem is equal to $ \Sigma^{-1}$. However, the number of available samples in many applications is  smaller than the dimension of $\Sigma$. This makes  $\Sigma$  ill-conditioned or even singular, which would lead to large or unbounded entries for  the optimal solution of \eqref{opt_inv}. Furthermore, although $\Sigma_*^{-1}$ is assumed to be sparse, a small difference between  $\Sigma_*$ and $\Sigma$  would potentially make $S^{\text{opt}}$ highly dense.
In an effort to address the aforementioned issues,  consider the $l_1$-regularized version of~\eqref{opt_inv}
\begin{equation}\label{opt1}
\underset{
	\begin{subarray}{c}
	S\in\mathbb{S}^d_+
	\end{subarray}
}{\text{minimize}} \ -\log\det(S)+\text{trace}(\Sigma S)+ \lambda\|S\|_*,
\end{equation}
where $\lambda > 0$ is a regularization coefficient. Let the optimal solution of~\eqref{opt1} be denoted by $S^{\text{opt}}$. This problem is known as Graphical Lasso (GL). The  term $\|S\|_*$ in the objective function is defined as the summation of the absolute values of the off-diagonal entries in $S$. This additional penalty acts as a surrogate for promoting  sparsity in the off-diagonal elements of $S$, while ensuring that the problem is well-defined even with a singular input $\Sigma$. 
%The sparsity pattern of $S^{\text{opt}}$ is then used as an estimate of that of the  matrix  $\Sigma_*^{-1}$. 

It is well-known that the GL is computationally prohibitive for large-scale problems. One way to circumvent the problem of solving the highly-complex GL is to simply threshold the sample covariance matrix in order to obtain a candidate structure for the sparsity pattern of the optimal solution to the GL. It has been shown in several real-world problems, including brain connectivity networks and topology identification of electrical circuits~\cite{Sojoudi14, Sojoudi16}, that the thresholding method can correctly identify the nonzero pattern of $S^{\text{opt}}$. Recently, we have shown that the sparsity structure of the thresholding method coincides with that of the GL~\cite{Salar17} under some conditions on the sample covariance matrix. Although these conditions are not easy to verify, it is shown that they are generically satisfied when a sparse solution for the GL is sought, or equivalently, when the regularization parameter in~\eqref{opt1} is large. Based on this observation,~\cite{Salar17} derives a closed-form solution for the GL problem that is globally or near-globally optimal for the GL, depending on the structure of the sample covariance matrix.

In this paper, we generalize the results of~\cite{Salar17} to the cases where the thresholded sample covariance matrix has a chordal structure. A matrix has a chordal structure if every cycle in its support graph with length of at least 4 has a chord. Clearly, this class of sparsity structures includes acyclic graphs. First, we revisit the conditions introduced in~\cite{Salar17} for the equivalence of the sparsity structures found using the GL and the simple method of thresholding. We show that the conditions for this equivalence can be significantly simplified when the support graph of the thresholded covariance matrix is chordal. Furthermore, we show that under some mild assumptions, these conditions are automatically satisfied as the size of the sample covariance matrix grows, provided that they possess sparse and chordal structures. In the second part of the paper, we generalize the closed-form solution of the GL with acyclic thresholded sample covariance matrix to those with chordal structures. More specifically, we show that $S^{\text{opt}}$ can be obtained using a closed-form recursive formula when the thresholded sample covariance matrix has a chordal structure. As it is pointed out in~\cite{Mazumdar12}, most of the numerical algorithms for solving the GL has the worst case complexity of at least $\mathcal{O}(d^4)$. We show that the recursive formula requires a number of iterations growing linearly in the size of the problem, and that the complexity of each iteration is dependent on the size of the maximum clique in the sparsity graph. Therefore, given the thresholded sample covariance matrix (which can be obtained while constructing the sample covariance matrix), the complexity of solving the GL for sparse and chordal structures reduces from $\mathcal{O}(d^4)$ to $\mathcal{O}(d)$. In fact, we show the graceful scalability of the proposed recursive method in large-scale problems. Specifically, we show that, on average, the proposed method outperforms the best known algorithm for solving the GL by at least a factor of 16 with the sample correlation sizes between $1220\times 1220$ and $29902\times 29902$.

The Graphical Lasso technique is commonly-used for estimating the inverse covariance matrices of Gaussian distributions. However, a similar learning method can be employed for data samples with more general underlying distributions. More precisely, the GL corresponds to the minimization of the $l_1$-regularized log-determinant Bregman divergence, which is a widely-used metric for measuring the distance between the true and estimated parameters of a problem~\cite{Martin11}. Therefore, the theoretical results developed in this paper are applicable to more general inference problems.

%\subsection{Related Works}

{\bf Notations:} $\mathbb{R}^d$, $\mathbb{S}^d$, $\mathbb{S}^d_+$, and $\mathbb{S}^d_{++}$ are used to denote the sets of $d\times 1$ real vectors, $d\times d$ symmetric matrices, $d\times d$ positive-semidefinite matrices, and $d\times d$ positive-definite matrices, respectively. The symbols $\mathrm{trace}(M)$ and $\log\det(M)$ refer to the trace and the logarithm of the determinant of the matrix $M$, respectively. The $i^{\text{th}}$ and $(i,j)^{\text{th}}$ entries of the vector $\mathbf{m}$ and matrix $M$ are denoted by $\mathbf{m}_i$ and $M_{ij}$, respectively. $I_d$ refers to an $d\times d$ identity matrix. The sign of a scalar $x$ is shown by $\text{sign}(x)$. For a set ${D}$, $|{D}|$ refers to its cardinality. The inequality $M\succ 0$ ($M\succeq 0$) means that $M$ is positive-(semi)definite. For a graph $\mathcal{G}$, $\mathcal{N}(k)$ denotes the set of neighbors of node $k$. Given a vector $\mathbf{m}\in\mathbb{R}^d$ and matrix $M\in\mathbb{S}^d$, define 
\begin{align}
	&\|M\|_* = \sum_{i=1}^{d}\sum_{j=1}^{d}|M_{ij}|-\sum_{i=1}^{d}|M_{ii}|\\
	&\|M\|_{\max} = \max_{i\not=j}|M_{ij}|\\
	&\|\mathbf{m}\|_{\max}=\max_{i}|\mathbf{m}_i|
\end{align}
An \emph{index set} is a sorted subset of the integers $\{1,2,\ldots,d\}$.
The number of elements in the index set $I$ is denoted by $|I|$. Given
index sets $I$ and $J$, we define
\[
(P_{IJ})_{i,j}=\begin{cases}
1 & I(i)=J(j),\\
0 & \text{otherwise.}
\end{cases}
\]
%in order to match the notation of Andersen, Dahl and Vandenberghe
%during our description of the max-det matrix completion algorithm.

A \emph{sparsity pattern} is a symmetric binary matrix. A $d\times d$
matrix $X$ (not necessarily symmetric) is said \emph{to have sparsity
	pattern} $E$ if $X_{i,j}=0$ whenever $E_{i,j}=0$; the set $\R_{E}^{d\times d}$
(resp. $\S_{E}^{d}$) refers to the $d\times d$ matrices (resp. $d\times d$
symmetric matrices) with sparsity pattern $E$. The \emph{Euclidean
	projection onto sparsity pattern} is denoted as $\Pi_{E}(\cdot)$:
the $(i,j)^{\text{th}}$ element of $\Pi_{E}(M)$ is zero if $E_{i,j}=0$,
and $M_{i,j}$ if $E_{i,j}=1$.

\section{Preliminaries}

In this section, we review the properties of sparse and chordal matrices and their connection to the max-det matrix completion problem.

\subsection{Sparse Cholesky factorization}

Consider solving a $d\times d$ symmetric positive definite linear
system 
\[
Sx=b
\]
by Gaussian elimination. The standard procedure comprises a \emph{factorization}
step, where $S$ is decomposed into the (unique) Cholesky factor matrices
$LDL^{T}$, in which $D$ is diagonal and $L$ is lower-triangular
with a unit diagonal, and a \emph{substitution }step, where the two
triangular systems of linear equations $Ly=b$ and $DL^{T}x=y$ are solved to yield $x$. 

In the case where $S$ is sparse, the Cholesky factor $L$ is often
also sparse. It is common to store the sparsity pattern of $L$ in
the \emph{compressed column storage} format: a set of indices $I_{1},\ldots,I_{d}\subseteq\{1,\ldots,d\}$
in which
\begin{align}
	I_{j} & =\{i\in\{1,\ldots,d\}:i>j,L_{i,j}\ne0\},\label{eq:defI}
\end{align}
encodes the locations of off-diagonal nonzeros in the $j^{\text{th}}$ column
of $L$. (The diagonal elements are not included because the matrix
$L$ has a unit diagonal by definition). 

After storage has been allocated and the sparsity structure determined,
the numerical values of $D$ and $L$ are computed using a sparse
Cholesky factorization algorithm. This requires the use of the associated
\emph{elimination tree} $\Tree=\{V,\Edges\}$, which is a rooted tree
(or forest) on $d$ vertices, with edges $\mathcal{E}=\{\{1,p(1)\},\ldots,\{d,p(d)\}\}$
defined to connect each $j^{\text{th}}$ vertex to its parent at the $p(j)^{\text{th}}$
vertex (except root nodes, which have ``0'' as their parent), as
in
\begin{equation}
	p(j)=\begin{cases}
		\min I_{j} & |I_{j}|>0,\\
		0 & |I_{j}|=0,
	\end{cases}\label{eq:elim_tree}
\end{equation}
in which $\min I_{j}$ indicates the (numerically) smallest index in
the index set $I_{j}$~\cite{Liu90}. The elimination tree encodes the dependency
information between different columns of $L$, thereby allowing information
to be passed without explicitly forming the matrix.

\subsection{Chordal sparsity patterns}

The \textit{support} or \textit{sparsity graph} of $E$, denoted by $\text{supp}(E)$, is defined as a graph with the vertex set $\mathcal{V} = \{1,2,...,d\}$ and the edge set $\mathcal{E}$, where $(i,j)\in\mathcal{E}$ if and only if $E_{ij}\not=0$ and $i\neq j$.
The pattern $E$ is said to be \emph{chordal} if its graph does not contain an induced cycle
with length greater than three. If $E$ is not chordal, then we may
add nonzeros to it until it becomes chordal; the resulting pattern
$E'$ is called a \emph{chordal completion} (or chordal embedding
or triangulation) of $E$. Any chordal completion $E'$ with at most
$O(d)$ nonzeros is a \emph{sparse} chordal completion of $E$.

A sparsity pattern $E$ is said to \emph{factor without fill} if every
$S$ with sparsity pattern $E$ can be factored into $LDL^{T}$ such
that $L+L^{T}$ also has sparsity pattern $E$. If a sparsity pattern
factors without fill, then it is chordal. Conversely, if a sparsity
pattern is chordal, then there exists a permutation matrix $Q$ such
that $QEQ^{T}$ factors without fill~\cite{Fulkerson65}. This permutation matrix $Q$
is called the \emph{perfect elimination ordering} of the chordal sparsity
pattern $E$.

\subsection{Recursive solution of the max-det problem.}

An important application of chordal sparsity is the
efficient solution of the maximum determinant matrix completion problem,
written
\begin{align}
	\hat{X}=\underset{X\succeq0}{\text{ maximize }} & \log\det(X)\label{eq:logdetp}\\
	\text{subject to } & X_{i,j}=C_{i,j}\text{ for all }C_{i,j}\ne0\nonumber 
\end{align}
for a given large-and-sparse matrix $C$ with sparsity pattern $E$. The optimal solution of the above optimization (when it exists) is called the max-det matrix completion of $C$, and is unique.
The Lagrangian dual of this problem is the following
\begin{align}
	\hat{S}=\underset{S\succeq0}{\text{ minimize }} & -\log\det S+\text{trace}(C S)+d,\label{eq:logdetd}\\
	\text{subject to } & S\in\S_{E}^{d},\nonumber 
\end{align}
with first-order optimality condition
\begin{equation}
	\Pi_{E}(\hat{S}^{-1})=C.\label{eq:foc}
\end{equation}
Strong duality gives a straightforward relation back to the primal
\begin{equation}
	\hat{X}=\hat{S}^{-1}.\label{eq:duality}
\end{equation}
Note that while $\hat{X}$ is (in general) a dense matrix, $\hat{S}$
is always sparse. Instead of attempting to solve the primal problem
(\ref{eq:logdetp}) for a dense matrix, we may opt to solve the dual
problem (\ref{eq:logdetd}) for a sparse matrix satisfying the optimality
condition (\ref{eq:foc}). In the case that the sparsity pattern $E$
factors without fill,~\cite{Dahl13} showed that
(\ref{eq:foc}) is actually a linear system of equations over the
Cholesky factor $L$ and $D$ of the solution $\hat{S}=LDL^{T}$;
their numerical values can be explicitly computed using a recursive formula. 

\vspace{2mm}
\begin{algor}(\cite{Dahl13}, Algorithm 4.2)\textbf{}\\
	\textbf{Input.} Matrix $C\in\S_{E}^{d}$ that has a positive definite
	completion.\textbf{}\\
	\textbf{Output.} The Cholesky factors $L$ and $D$ of $\hat{S}=LDL^{T}\in\S_{E}^{d}$
	that satisfy $\Pi_{E}(\hat{S}^{-1})=C$.\textbf{}\\
	\textbf{Algorithm.} Iterate over $j\in\{1,2,\ldots,d\}$ in reverse,
	i.e. starting from $d$ and ending in $1$. For each $j$, compute
	$D_{jj}$ and the $j^{\text{th}}$ column of $L$ from
	\begin{align}
		& L_{I_{j}j}=-V_{j}^{-1}S_{I_{j}j},\nonumber\\ 
		& D_{jj}=(C_{jj}+C_{I_{j}j}^{T}L_{I_{j}j})^{-1}\nonumber
	\end{align}
	and compute the update matrices
	\[
	V_{i}=P_{J_{j}I_{i}}^{T}\begin{bmatrix}C_{jj} & C_{I_{j}j}^{T}\\
	C_{I_{j}j} & V_{j}
	\end{bmatrix}P_{J_{j}I_{i}}
	\]
	for each $i$ satisfying $p(i)=j$, i.e. each child of $j$ in the
	elimination tree.\label{alg1}\end{algor}

\vspace{2mm}
Of course, if the sparsity pattern $E$ is chordal, then we may find
the perfect elimination ordering $Q$ in linear time~\cite{Lieven15}, and apply the above algorithm
to the matrix $QCQ^{T}$, whose sparsity pattern does indeed factor
without fill.

The algorithm takes $d$ steps, and the $j^{\text{th}}$ step requires a size-$|I_{j}|$
linear solve and vector-vector product. The \textit{treewidth} of the sparsity graph is defined as $w = w(\text{supp}(E)) = \max_j |I_j| -1, $ and has the interpretation of the largest clique in $\text{supp}(E)$ minus one. Combined, the algorithm has
time complexity $\mathcal{O}(w^{3}d)$. This
means that the matrix completion algorithm is linear-time if the treewidth of $\text{supp}(E)$ is in the order of $\mathcal{O}(1)$.

\section{Main Results}
To streamline the presentation, with no loss of generality we assume that $\Sigma$ used in~\eqref{opt1} is the sample correlation matrix. This means that the diagonal elements of $\Sigma$ are normalized to 1 and the off-diagonal elements are between $-1$ and $1$. The results of this paper can readily be generalized to an arbitrary sample covariance matrix after appropriate rescaling. The following definitions are borrowed from~\cite{Salar17}.
%\begin{definition}
%	Given a  matrix $M\in \mathbb{S}^d$, the {\bf support or sparsity graph} of $M$ denoted by $\text{supp}(M)$ is defined as a graph with the vertex set $\mathcal{V} = \{1,2,...,d\}$ and the edge set $\mathcal{E}$, where $(i,j)\in\mathcal{E}$ if and only if $M_{ij}\not=0$ and $i\neq j$.
%\end{definition}
\begin{definition} \label{def:dd1}
	
	A  matrix $M\in\mathbb S^d$ is called {\bf inverse-consistent} if there exists a matrix $N\in\mathbb S^d$ with zero diagonal elements such that
	\begin{subequations}
		\begin{align}
		&M+N\succ 0\\
		&\mathrm{supp}(N)\subseteq\left(\mathrm{supp}(M)\right)^{(c)}\\
		&  \mathrm{supp}\left((M+N)^{-1})\right)\subseteq\mathrm{supp}(M),
		\end{align}
	\end{subequations}
	where $(\mathrm{supp}(M))^{(c)}$ is the complement of $\mathrm{supp}(M)$. The matrix $N$ is called an {\bf inverse-consistent complement } of $M$ and is denoted as $M^{(c)}$.
	
	Moreover, $M$ is called {\bf sign-consistent} if the $(i,j)$ entries of $M$ and $(M+M^{(c)})^{-1}$ are nonzero and have opposite signs for every $(i,j)\in\mathrm{supp}(M)$.
	
\end{definition}

\vspace{2mm}
\begin{example} {\rm Consider the matrix:
		\begin{equation}
		M=\left[\begin{array}{cccc}
		1& 0.3 &0  & 0\\
		0.3 & 1 & -0.4 &0\\
		0 & -0.4 & 1 & 0.2\\
		0 & 0 & 0.2 & 1
		\end{array}\right]
		\end{equation}
		We show that $M$ is both inverse- and sign-consistent.
		Consider the matrix $M^{(c)}$ defined as
		\begin{equation}
		M^{(c)}=\left[\begin{array}{cccc}
		0& 0 &-0.12  & -0.024\\
		0& 0& 0 &-0.08\\
		-0.12 & 0 & 0 & 0\\
		-0.024 & -0.08 & 0 &0
		\end{array}\right]
		\end{equation}
		$(M+M^{(c)})^{-1}$ can be written as
		\begin{equation}
		\left[\begin{array}{cccc}
		\frac{1}{0.91}& \frac{-0.3}{0.91} &0  & 0\\
		\frac{-0.3}{0.91} & 1+\frac{0.09}{0.91}+\frac{0.16}{0.84} & \frac{0.4}{0.84} &0\\
		0 & \frac{0.4}{0.84} & 1+\frac{0.16}{0.84}+\frac{0.04}{0.96}  & \frac{-0.2}{0.96}\\
		0 & 0 & \frac{-0.2}{0.96} & \frac{1}{0.96}
		\end{array}\right]
		\end{equation}
		Note that: 
		\begin{itemize}
			\item $M+M^{(c)}$ is positive-definite.
			\item The sparsity graph of $M$ is the complement of that of $M^{(c)}$.
			\item The sparsity graphs of $M$ and $(M+M^{(c)})^{-1}$ are equivalent.
			\item  The nonzero off-diagonal entries of $M$ and $(M+M^{(c)})^{-1}$ have opposite signs. 
		\end{itemize}
	Therefore, it can be inferred that $M$ is both  inverse- and sign-consistent, and $M^{(c)}$ is its inverse-consistent complement.}
\end{example}

\vspace{2mm}
In \cite{Salar17}, it has been shown that every positive definite matrix has a unique inverse-consistent complement.

\vspace{2mm}
\begin{definition}
	\label{def:dd3}
	Given a graph $\mathcal G$ and a scalar $\alpha$, define  $\beta(\mathcal G,\alpha)$ as the maximum of $\|M^{(c)}\|_{\max}$ over all inverse-consistent positive-definite matrices $M$ with the diagonal entries equal to 1  such that $\mathrm{supp}(M)=\mathcal G$ and  $\|M\|_{\max}\leq \alpha$. 
\end{definition}

\vspace{2mm}
Without loss of generality and due to the non-singularity of $\Sigma_*$, one can assume that all elements of $\Sigma$ are nonzero. Let $\sigma_1, \sigma_2,...,\sigma_{d(d-1)/2}$ be the sorted upper triangular entries of $\Sigma$ such that 
\begin{equation}
	\sigma_1> \sigma_2>...>\sigma_{d(d-1)/2}>0
\end{equation}

\begin{definition}
	Define the {\bf residue of $\Sigma$ at level $k$ relative to $\lambda$ } as a matrix $\Sigma^{\mathrm{res}}(k,\lambda)\in\mathbb S^d$ whose $(i,j)^{\text{th}}$ entry is equal to 
	$
	\Sigma_{ij}-\lambda\times \mathrm{sign}(\Sigma_{ij})
	$
	if $|\Sigma_{ij}|>\lambda$, and  equal to 0 otherwise.
\end{definition}

\vspace{2mm}
Notice that $\Sigma^{\mathrm{res}}(k,\lambda)$ is the soft-thresholded sample correlation matrix with threshold $\lambda$. For simplicity of notation, we omit the arguments $k$ and $\lambda$ in $\Sigma^{\mathrm{res}}(k,\lambda)$ whenever the equivalence is implied from the context.
The following theorem is borrowed from~\cite{Salar17}.

\vspace{2mm}
\begin{theorem} \label{thm:tt1}
	The thresholding method and the GL have the same sparsity patterns if the following conditions are satisfied for $\lambda\in(\sigma_{k+1},\sigma_k)$: 
	\begin{itemize}
		\item {\bf Condition 1-i:} $I_d+\Sigma^{\mathrm{res}}$ is positive definite.
		\item {\bf Condition 1-ii:} $I_d+\Sigma^{\mathrm{res}}$ is sign-consistent.
		\item {\bf Condition 1-iii:} The relation
		\begin{equation}
		\beta\left(\text{supp}(\Sigma^{\mathrm{res}}),\sigma_1-\lambda\right)\leq \lambda-\sigma_{k+1}.
		\end{equation}~\hfill$\blacksquare$
	\end{itemize}
\end{theorem}

\vspace{2mm}
In~\cite{Salar17}, it is pointed out that the aforementioned conditions in Theorem~\ref{thm:tt1} are expected to hold if a sparse solution for the GL problem is sought. However, efficient verification of these conditions is yet to be addressed in practice. It has been observed that the last condition plays the most important role in verifying the optimality conditions for the sparsity pattern of the thresholded sample correlation matrix. 

\subsection{Upper Bound for $\beta(\mathcal{G},\alpha)$ in Chordal Graphs}

In what follows, we derive an upper bound on $\beta(\mathcal{G},\alpha)$ for chordal graphs and show that under some mild assumptions, Condition 1-iii is satisfied as the size of the sample correlation matrix grows.

\vspace{2mm}
\begin{theorem}\label{thm3}
	Suppose that the following conditions hold:
	\begin{itemize}
		\item{\bf Condition 2-i:} $\mathcal{G}$ is chordal.
		\item{\bf Condition 2-ii:} ${w}(\mathcal{G})\leq\frac{2}{3}(d-1)$.
		\item{\bf Condition 2-iii:} $\alpha<\frac{1}{{w}(\mathcal{G})\sqrt{d-{w}(\mathcal{G})-1}+{w}(\mathcal{G})-1}$
	\end{itemize}
	Then, we have
	\begin{equation}\label{upper}
	\beta(\mathcal{G},\alpha)\leq\frac{{w}(\mathcal{G})\sqrt{d-{w}(\mathcal{G})-1}\times\alpha^2}{1-(w(\mathcal{G})-1)\times\alpha}
	\end{equation}
\end{theorem}

\vspace{2mm}
\begin{proof}
	The proof is provided in Appendix.
\end{proof}

\vspace{2mm}

Conditions 2-ii and 2-iii are guaranteed to be satisfied for small values of $\alpha$. In such circumstances, chordal structure for $\mathcal{G}$ is enough to verify the validity of~\eqref{upper}. Based on this theorem, the next corollary shows that the Condition 1-iii in Theorem~\ref{thm:tt1} is guaranteed to be satisfied when the support graph of the sample correlation matrix is sparse and large enough.
\vspace{2mm}
\begin{corollary}\label{cor1}
	Suppose that the following conditions hold for some $\delta\geq 0$ and $\epsilon>1/2$:
	\begin{subequations}
		\begin{align}
		& \lambda-\sigma_{k+1} = \mathcal{O}(d^{-\epsilon})\label{sigmak1}\\
		& \sigma_1-\lambda = \mathcal{O}({d^{-(\epsilon-\delta)}})\label{sigma1}\\
		& w(\mathcal{G}) = \mathcal{O}(1)\label{tw}\\
		& \delta< \epsilon/2-1/4\label{upperep}
		\end{align}
	\end{subequations}
	Then, there exists a $d_0>0$ such that for every $d\geq d_0$, the Condition 1-iii in Theorem~\ref{thm:tt1} is satisfied.
\end{corollary}

\vspace{2mm}
\begin{proof}
	The proof is provided in Appendix.
\end{proof}

\vspace{2mm}
Corollary~\ref{cor1} implies that, if the sample correlation matrix is large enough and the rate of decrease in $\sigma_1-\lambda$, as a function of the dimension of the data, is not much smaller than that of $\lambda-\sigma_{k+1}$, then, Condition 1-iii is automatically satisfied. For instance, suppose that $\epsilon = 2$ in~\eqref{sigmak1}. Then, Corollary~\ref{cor1} implies that for large enough values of $d$, Condition 1-iii is satisfied if $w(\mathcal{G}) = \mathcal{O}(1)$ and there exists a $\gamma>0$ such that $\sigma_1-\lambda = \mathcal{O}(d^{-1.25-\gamma})$.

\begin{remark}
	Although~\eqref{upper} only holds for chordal graphs, one can generalize Theorem~\ref{thm3} to non-chordal graphs, under some additional assumptions. In particular, suppose that $\tilde{\mathcal{G}}$ is the chordal completion of a non-chordal graph $\mathcal{G}$. Then, one can easily verify that~\eqref{upper} holds for $\beta(\mathcal{G},\alpha)$ if $\beta(\mathcal{G},\alpha)\leq\beta(\tilde{\mathcal{G}},\alpha)$. Indeed, we could show that the monotonic behavior of $\beta(\cdot, \cdot)$ is maintained under fairly general conditions. Due to the space restrictions, this generalization is not included in this paper.
\end{remark}
\subsection{Max-Det Matrix Completion for Graphical Lasso}

In this subsection, we show that if the equivalence between the thresholding method and GL holds, the optimal solution of the GL can be obtained using Algorithm~\ref{alg1}.

%\begin{definition}
%	Given a positive-definite matrix $M$, $\bar{M}$ is called the {\bf max-det positive-definite matrix completion} of $M$ if it is the unique maximizer of 
%	\begin{subequations}
%		\begin{align}
%		\underset{
%			\begin{subarray}{c}
%			X\in\mathbb{S}^d_+
%			\end{subarray}
%		}{\text{maximize}}& \ \log\det(X)\\
%		\text{s.t.} & \ \ X_{ij} = M_{ij}\ \ \text{if}\ M_{ij}\not= 0
%		\end{align}
%	\end{subequations}
%For simplicity of notation, $\bar{M}$ is called the max-det completion of $M$. 
%\end{definition}

%In the next theorem, we show that if the thresholding method finds the optimal sparsity structure of the solution in GL, then the GL can be reformulated as a max-(log)-det optimization problem with sparsity constraints.

\vspace{2mm}
\begin{theorem}\label{thm2}
	Assume that the thresholded sample correlation matrix has the same sparsity pattern as the optimal solution of the GL and $\lambda\geq 0.5$. Then, 
	\begin{itemize}
		\item[1.] $(S^{\text{opt}})^{-1}$ is the unique max-det completion of $\Sigma^{\mathrm{res}}$.
		\item[2.] Algorithm~\ref{alg1} can be used to find $S^{\text{opt}}$ if $\mathrm{supp}(\Sigma^{\mathrm{res}})$ is chordal.
	\end{itemize} 
\end{theorem}

\vspace{2mm}
\begin{proof}
	The proof is provided in Appendix.
\end{proof}

\vspace{2mm}
Recall that the main goal of the GL is to promote the sparsity structure of the inverse correlation matrix. In order to obtain a sparse solution, the regularization coefficient should be large, relative to the absolute values of the off-diagonal elements in the sample correlation matrix. Under such circumstances, the conditions delineated in Theorem~\ref{thm:tt1} are satisfied and the sparsity pattern of the simple thresholding technique corresponds to that of the GL. Theorem~\ref{thm2} uses this result to show that for large values of $\lambda$, instead of merely identifying its sparsity structure, the thresholded sample correlation matrix can be further exploited to find the optimal solution of the GL problem by solving its corresponding max-det matrix completion problem. Note that the first part of Theorem~\ref{thm2} is independent of the structure of the thresholded sample correlation matrix. However, the second part of the theorem suggests that solving its corresponding max-det matrix completion problem can be much easier than the GL problem and can be performed in linear time using Algorithm~\ref{alg1} when the thresholded sample correlation matrix has a sparse and chordal structure.

While the focus of this paper is on the thresholded sample correlation matrices with chordal structures, the presented method may be extended to matrices with non-chordal sparsity patterns. Note that for non-chordal structures, the provided recursive formula does not necessarily result in the optimal solution. However, it has been shown in~\cite{Dahl08} that efficient implementations of Newton and conjugate gradient methods for max-det matrix completion problem are possible when the sparsity structure of the problem has a sparse chordal completion. The detailed analysis of this extension is left as future work. Furthermore, as it is pointed out in \cite{Mazumdar12} and \cite{Witten11}, the disjoint components of the sparsity graph induced by the thresholding of sample correlation matrix can be treated independently since the GL is decomposed into multiple smaller size problems over these disjoint components. Therefore, the proposed method can be applied to every chordal component even if the overall sparsity graph of the thresholded sample correlation matrix does not benefit from a chordal structure.

\section{Numerical Results}

\begin{table*}
	\begin{center}
		\begin{tabular}{ c||c|c||c||c|c|c||c|c|c}
			& & & recursive & \multicolumn{3}{|c||}{\texttt{QUIC}} & \multicolumn{3}{|c}{\texttt{GLASSO}}\\
			\hline
			test case & matrix size & max clique size & run. time & run. time & opt. gap & speedup & run. time & opt. gap & speedup\\ 
			\hline
			{fpga-dcop-01}$^+$ & 1220 & 10 & 0.17 & 0.46 & $<10^{-7}$ & 2.71 & 45.93 & $<10^{-7}$ & 99.85\\
			\hline
			west1505$^+$ & 1505 & 338 & 0.89 & 9.56 & $<10^{-5}$ & 10.74 & 137.59 & $<10^{-5}$ & 154.60\\
			\hline
			netscience$^+$ & 1589 & 20 & 0.21 & 0.91 & $<10^{-7}$ & 4.33 & 108.04 & $<10^{-7}$ & 528.53\\
			\hline
			lung1$^+$ & 1650 & 4 & 0.24 & 1.70 & $<10^{-7}$ & 7.08 & 93.63 & $<10^{-7}$ & 368.50\\
			\hline
			cryg2500$^+$ & 2500 & 75 & 0.51 & 6.39 & $<10^{-7}$ & 12.53 & 446.23 & $<10^{-7}$ & 892.47\\
			\hline
			freeFlyingRobot-7$^+$ & 3918 & 35 & 0.86 & 18.13 & $<10^{-8}$ & 21.08 & 2066.80 & $<10^{-8}$ & 2319.98\\
			\hline
			freeFlyingRobot-14$^+$ & 5985 & 35 & 1.53 & 40.54 & $<10^{-8}$ & 26.50 & $*$&$*$ &$*$\\
			\hline
			CASE13659PEGASE$^\dagger$ & 13659 & 35 & 5.34 & 260.04 & $<10^{-8}$ & 48.70& $*$&$*$ &$*$\\
			\hline
			OPF-6000$^+$ & 29902 & 52 & 78.97 & $*$ & $*$ & $*$& $*$&$*$ &$*$\\
			
		\end{tabular}
	\end{center}
	\caption{ \footnotesize The running time, relative optimality gap, and the speedup of the proposed recursive formula, compared to the \texttt{GLASSO} and \texttt{QUIC} algorithms. The superscripts ``$^+$'' and ``$^\dagger$'' correspond to the test cases chosen from SuiteSparse Matrix Collection and MATPOWER package, respectively.}\label{Polish}
\end{table*}

Using the method proposed in this paper, we solve the GL problem on various large-scale problems whose thresholded sample correlation matrices have chordal structures. All the test cases are collected from the \textit{SuiteSparse Matrix Collection}~\cite{Sparse11} and \textit{MATPOWER package} ~\cite{Zimmerman11, Coffrin141}. These are publicly available and widely-used datasets for large-and-sparse matrices from real-world problems. The simulations are run on a laptop computer with an Intel Core i7 quad-core 2.50 GHz CPU and 16GB RAM. The results reported in this section are for a serial implementation in MATLAB. 
\subsection{Data Generation}
For each test case, we take the following steps to design the sample correlation matrix. 
\begin{itemize}
	\item[1.] First, the nonzero structure of the matrix for a given test case is exploited and a Symbolic Cholesky Factorization is performed to arrive at a chordal embedding of the given structure~\cite{Agrawal93}. In other words, we augment the sparsity graph corresponding to the considered test case with additional edges to obtain a sparse chordal completion of the graph.
	\item[2.] The elements of the sample correlation matrix corresponding to the edges in the extended graph are chosen randomly from the union of the intervals $[-0.50,-0.55]$ and $[0.50,0.55]$. The rest of the elements are randomly chosen from the interval $[-0.20, 0.20]$.
	\item[3.] The diagonal elements of the sample correlation matrix are elevated according to the off-diagonal elements in order to make the sample correlation matrix positive semidefinite. The resulted matrix is normalized, if necessary.
\end{itemize}

\subsection{Discussion}

We consider different test cases corresponding to various real-world problems in materials science, power networks, circuit simulation, optimal control, fluid dynamics, social networks, and chemical process simulation. The size of the variable matrices in the GL problem for the investigated problems ranges from $1220\times 1220$ (with approximately 700 thousands variable elements) to $29902\times 29902$ (with approximately 447 million variable elements). We compare the running time and objective function of our proposed method with two other state-of-the-art algorithms, namely the \texttt{GLASSO}~\cite{Friedman08} and \texttt{QUIC}~\cite{Hsieh14} algorithms (downloaded from \href{url}{http://statweb.stanford.edu/$\sim$tibs/glasso/} and \href{url}{http://bigdata.ices.utexas.edu/software/1035/}, respectively). The GLASSO algorithm is the most widely-used algorithm for the GL, while the QUIC algorithm is commonly regarded as the fastest solver for the GL. Define the \textit{relative optimality gap} as the normalized difference between the objective functions of the proposed solution and the solutions that are obtained by the other two methods. We consider a 2-hour time limit for the solvers. 

Table~\ref{Polish} shows the results of our simulations. It can be observed that the proposed recursive method significantly outperforms the \texttt{GLASSO} and \texttt{QUIC} algorithms in terms of running time, while achieving a negligible relative optimality gap in most of the test cases. More specifically, for the first 8 cases, the proposed recursive method is $16.70$ times faster than the \texttt{QUIC}. For the largest test case, \texttt{QUIC} does not find the optimal solution within the 2-hour time limit while the proposed recursive formula obtains the optimal solution in less than 2 minutes. Furthermore, the recursive method is 726 times faster than \texttt{GLASSO} algorithm for the first 6 test cases. However, this algorithm does not find the optimal solution for the 3 largest test cases within the 2-hour time limit.

\section{Conclusions}
In many graphical learning problems, the goal is to obtain a sparsity graph that describes the conditional independence of different elements in the available dataset via sparse inverse covariance estimation. The Graphical Lasso (GL) is one of the most commonly-used methods for addressing this problem.  It is known that, in high-dimensional settings, the GL is computationally-prohibitive. A cheap alternative method for finding the sparsity pattern of the inverse covariance matrix is a simple thresholding method performed on the sample covariance matrix of the data. Recently, we have provided sufficient conditions under which the thresholding is equivalent to the GL in terms of the sparsity pattern of the graphical model. Based on this result, we have shown that the GL has a closed-form solution when the thresholded sample covariance matrix is acyclic. In this paper, this result is generalized to the problems where the thresholding results in a chordal structure. It is shown that the sufficient conditions for the equivalence of the thresholding and GL can be significantly simplified for chordal structures and is expected to hold as the dimension of the data increases. Furthermore, it is shown that the GL can be reduced to a maximum determinant matrix completion problem when the thresholding is equivalent to the GL, and for chordal structures, the corresponding matrix completion problem has a simple recursive formula. The performance of the derived recursive formula is compared with the other commonly-used methods and shown that, for the large-scale GL problems, the proposed method significantly outperforms other methods in terms of their running times.

	\medskip
	
	\bibliographystyle{IEEEtran}
	\bibliography{reference_r}

\begin{thebibliography}{10}
\providecommand{\url}[1]{#1}
\csname url@samestyle\endcsname
\providecommand{\newblock}{\relax}
\providecommand{\bibinfo}[2]{#2}
\providecommand{\BIBentrySTDinterwordspacing}{\spaceskip=0pt\relax}
\providecommand{\BIBentryALTinterwordstretchfactor}{4}
\providecommand{\BIBentryALTinterwordspacing}{\spaceskip=\fontdimen2\font plus
\BIBentryALTinterwordstretchfactor\fontdimen3\font minus
  \fontdimen4\font\relax}
\providecommand{\BIBforeignlanguage}[2]{{%
\expandafter\ifx\csname l@#1\endcsname\relax
\typeout{** WARNING: IEEEtran.bst: No hyphenation pattern has been}%
\typeout{** loaded for the language `#1'. Using the pattern for}%
\typeout{** the default language instead.}%
\else
\language=\csname l@#1\endcsname
\fi
#2}}
\providecommand{\BIBdecl}{\relax}
\BIBdecl

\bibitem{Garcke01}
J.~Garcke, M.~Griebel, and M.~Thess, ``Data mining with sparse grids,''
  \emph{Computing}, vol.~67, no.~3, pp. 225--253, 2001.

\bibitem{Wright10}
J.~Wright, Y.~Ma, J.~Mairal, G.~Sapiro, T.~S. Huang, and S.~Yan, ``Sparse
  representation for computer vision and pattern recognition,''
  \emph{Proceedings of the IEEE}, vol.~98, no.~6, pp. 1031--1044, 2010.

\bibitem{Sojoudi14}
S.~Sojoudi and J.~Doyle, ``Study of the brain functional network using
  synthetic data,'' \emph{52nd Annual Allerton Conference on Communication,
  Control, and Computing (Allerton)}, pp. 350--357, 2014.

\bibitem{Fardad11}
M.~Fardad, F.~Lin, and M.~R. Jovanovi{\'c}, ``Sparsity-promoting optimal
  control for a class of distributed systems,'' \emph{American Control
  Conference}, pp. 2050--2055, 2011.

\bibitem{SODC2016}
S.~Fattahi and J.~Lavaei, ``On the convexity of optimal decentralized control
  problem and sparsity path,'' \emph{American Control Conference}, 2017.

\bibitem{Salar17}
S.~Fattahi and S.~Sojoudi, ``Graphical lasso and thresholding: Equivalence and
  closed-form solutions,'' \emph{\url{https://arxiv.org/abs/1708.09479}}, 2017.

\bibitem{Candes07}
E.~Candes and J.~Romberg, ``Sparsity and incoherence in compressive sampling,''
  \emph{Inverse Problems}, vol.~23, no.~3, pp. 969--985, 2007.

\bibitem{Friedman08}
J.~Friedman, T.~Hastie, and R.~Tibshirani, ``Sparse inverse covariance
  estimation with the graphical lasso,'' \emph{Biostatistics}, vol.~9, no.~3,
  pp. 432--441, 2008.

\bibitem{Banerjee08}
O.~Banerjee, L.~E. Ghaoui, and A.~d'Aspremont, ``Model selection through sparse
  maximum likelihood estimation for multivariate {Gaussian} or binary data,''
  \emph{Journal of Machine learning research}, vol.~9, pp. 485--516, 2008.

\bibitem{Sojoudi16}
S.~Sojoudi, ``Equivalence of graphical lasso and thresholding for sparse
  graphs,'' \emph{Journal of Machine Learning Research}, vol.~17, no. 115, pp.
  1--21, 2016.

\bibitem{Mazumdar12}
R.~Mazumder and T.~Hastie, ``Exact covariance thresholding into connected
  components for large-scale graphical lasso,'' \emph{Journal of Machine
  Learning Research}, vol.~13, pp. 781--794, 2012.

\bibitem{Witten11}
D.~M. Witten, J.~H. Friedman, and N.~Simon, ``New insights and faster
  computations for the graphical lasso,'' \emph{Journal of Computational and
  Graphical Statistics}, vol.~20, no.~4, pp. 892--900, 2011.

\bibitem{Martin11}
P.~Ravikumar, M.~J. Wainwright, G.~Raskutti, and B.~Yu, ``High-dimensional
  covariance estimation by minimizing $l_1$-penalized log-determinant
  divergence,'' \emph{Electronic Journal of Statistics}, vol.~5, pp. 935--980,
  2011.

\bibitem{Liu90}
J.~W. Liu, ``The role of elimination trees in sparse factorization,''
  \emph{SIAM Journal on Matrix Analysis and Applications}, vol.~11, no.~1, pp.
  134--172, 1990.

\bibitem{Fulkerson65}
D.~Fulkerson and O.~Gross, ``Incidence matrices and interval graphs,''
  \emph{Pacific journal of mathematics}, vol.~15, no.~3, pp. 835--855, 1965.

\bibitem{Dahl13}
J.~D. Andersen, Martin~S. and L.~Vandenberghe, ``Logarithmic barriers for
  sparse matrix cones,'' \emph{Optimization Methods and Software}, vol.~28,
  no.~3, pp. 396--423, 2013.

\bibitem{Lieven15}
L.~Vandenberghe and M.~S. Andersen, ``Chordal graphs and semidefinite
  optimization,'' \emph{Foundations and Trends$^{\textregistered}$ in
  Optimization}, vol.~1, no.~4, pp. 241--433, 2015.

\bibitem{Dahl08}
L.~V. Dahl, Joachim and V.~Roychowdhury, ``Covariance selection for nonchordal
  graphs via chordal embedding,'' \emph{Optimization Methods $\&$ Software},
  vol.~23, no.~4, pp. 501--520, 2008.

\bibitem{Sparse11}
T.~A. Davis and Y.~Hu, ``The university of florida sparse matrix collection,''
  \emph{ACM Transactions on Mathematical Software (TOMS)}, vol.~38, no.~1,
  p.~1, 2011.

\bibitem{Zimmerman11}
R.~D. Zimmerman, C.~E. Murillo-S{\'a}nchez, and R.~J. Thomas, ``Matpower:
  Steady-state operations, planning, and analysis tools for power systems
  research and education,'' \emph{IEEE Transactions on power systems}, vol.~26,
  no.~1, pp. 12--19, 2011.

\bibitem{Coffrin141}
C.~Coffrin, D.~Gordon, and P.~Scott, ``Nesta, the nicta energy system test case
  archive,'' \emph{arXiv preprint arXiv:1411.0359}, 2014.

\bibitem{Agrawal93}
A.~Agrawal, P.~Klein, and R.~Ravi, ``Cutting down on fill using nested
  dissection: provably good elimination orderings,'' \emph{Graph Theory and
  Sparse Matrix Computation, Springer}, pp. 31--55, 1993.

\bibitem{Hsieh14}
C.~J. Hsieh, M.~A. Sustik, I.~S. Dhillon, and P.~Ravikumar, ``Quic: quadratic
  approximation for sparse inverse covariance estimation,'' \emph{Journal of
  Machine Learning Research}, vol.~15, no.~1, pp. 2911--2947, 2014.

\bibitem{Fukuda01}
M.~K. K.~M. Fukuda, Mituhiro and K.~Nakata, ``Exploiting sparsity in
  semidefinite programming via matrix completion i: General framework,''
  \emph{SIAM Journal on Optimization}, vol.~11, no.~3, pp. 647--674, 2001.

\end{thebibliography}
	
	\appendix
	
	First, we present the KKT conditions for the optimal solution of~\eqref{opt1}.
	\begin{lemma}\label{expsol} 
		A matrix $S^{\text{opt}}$ is the optimal solution of the GL if and only if it satisfies the following conditions for every $i,j\in\{1,2,...,d\}$:
		\begin{subequations}\label{optcon}
			\begin{align}
			& (S^{\text{opt}})^{-1}_{ij} = \Sigma_{ij}&& \text{if}\quad i=j\\
			& (S^{\text{opt}})^{-1}_{ij} = \Sigma_{ij}+\lambda\times \text{\rm sign}(S^{\text{opt}}_{ij})&& \text{if}\quad S^{\text{opt}}_{ij}\not=0\label{con2}\\
			& \Sigma_{ij}-\lambda\leq (S^{\text{opt}})^{-1}_{ij} \leq \Sigma_{ij}+\lambda&& \text{if}\quad S^{\text{opt}}_{ij}=0
			\end{align}
		\end{subequations}
		where $(S^{\text{opt}})^{-1}_{ij}$ denotes the $(i,j)^{\text{th}}$ entry of $(S^{\text{opt}})^{-1}$.
	\end{lemma}

\vspace{2mm}
\begin{proof}
	The proof is straightforward and omitted for brevity.
\end{proof}

\vspace{2mm}
Next, we provide the proof of Theorem~\ref{thm2} based on Lemma~\ref{expsol}.

\vspace{2mm}
	\textit{Proof of Theorem~\ref{thm2}:} The second part of the theorem is a direct consequence of the first part and Algorithm~\ref{alg1}. To prove the first part, consider $(i,j)$ such that $\Sigma^{\mathrm{res}}_{ij}\not= 0$. Due to the assumption, the sparsity structure of $\Sigma^{\mathrm{res}}$ corresponds to that of $S^{\text{opt}}$. Therefore, we have $\Sigma^{\mathrm{res}}_{ij}\not= 0$. First, we show that $(S^{\text{opt}})^{-1}_{ij} = \Sigma_{ij}-\lambda\times \text{\rm sign}(\Sigma_{ij})$. By contradiction, suppose $(S^{\text{opt}})^{-1}_{ij} \not= \Sigma_{ij}-\lambda\times \text{\rm sign}(\Sigma_{ij})$. Since $S^{\text{opt}}_{ij}\not=0$, we must have $(S^{\text{opt}})^{-1}_{ij} = \Sigma_{ij}+\lambda\times \text{\rm sign}(\Sigma_{ij})$ based on~\eqref{con2}. Due to positive-definiteness of $(S^{\text{opt}})^{-1}$, one can write
	\begin{equation}\label{eqpos}
	1-(\Sigma_{ij}+\lambda\times \text{\rm sign}(\Sigma_{ij}))^2 > 0
	\end{equation}
	However, since $\Sigma^{\mathrm{res}}_{ij}\not= 0$, we have $|\Sigma_{ij}|>\lambda$. This implies 
	\begin{equation}\label{eqpos2}
	1-(2\lambda)^2\geq 1-(\Sigma_{ij}+\lambda\times \text{\rm sign}(\Sigma_{ij}))^2
	\end{equation}
	\eqref{eqpos} and \eqref{eqpos2} entails $\lambda < 0.5$ which is a contradiction. Therefore, one can invoke Theorem 2.4 in~\cite{Fukuda01} to show that $(S^{\text{opt}})^{-1}$ is indeed the max-det completion of $\Sigma^{\mathrm{res}}$ based on the following relations:
	\begin{itemize}
		\item $(S^{\text{opt}})^{-1}_{ij} = \Sigma^{\mathrm{res}}_{ij}$ if $\Sigma^{\mathrm{res}}_{ij}\not=0$.
		\item $S^{\text{opt}}_{ij} = 0$ if $\Sigma^{\mathrm{res}}_{ij}=0$.
	\end{itemize}
	This completes the proof.~\hfill$\blacksquare$
	
	\vspace{2mm}
	
	Given $M\in\mathbb{S}^d_{++}$, denote its unique max-det completion as $\bar{M}$. Furthermore, let $\bar{M}^{k}$ be \textit{the $k^{\text{th}}$ order max-det completion} of $X$ which is defined as the unique max-det completion of the submatrix corresponding to the last $d-k+1$ rows and columns of $M$. For simplicity, we abuse the notation and let $\bar{M}^{k}_{IJ}$ denote the submatrix of $\bar{M}^k$ whose rows and columns are indexed by $I$ and $J$ in $\bar{M}$.
	The following lemmas are crucial in order to prove Theorem~\ref{thm3}.

\vspace{2mm}
\begin{lemma}\label{lemma22}
	Condition 2-iii in Theorem~\ref{thm3} implies
	\begin{equation}
		\frac{{w}(\mathcal{G})\sqrt{d-{w}(\mathcal{G})-1}\times\alpha^2}{1-(w(\mathcal{G})-1)\times\alpha}< \alpha
	\end{equation}
\end{lemma}	

\vspace{2mm}
\begin{proof}
	Based on Condition 2-iii in Theorem~\ref{thm3}, one can write
	\begin{equation}\label{eq_alpha}
		\left({w}(\mathcal{G})\sqrt{d-{w}(\mathcal{G})-1}\right)\times\alpha^2<\alpha-\left({w}(\mathcal{G})-1\right)\times\alpha^2
	\end{equation}
	Furthermore, this condition yields $\alpha<1/({w}(\mathcal{G})-1)$. Dividing both sides of~\eqref{eq_alpha} by $1-({w}(\mathcal{G})-1)\times\alpha$ completes the proof.
\end{proof}

\vspace{2mm}
	\begin{lemma}\label{lemma2}
		Consider a partitioning of $\{1,2,...,d\}$ into 3 disjoint subsets $A$, $B$, and $C$. Given a positive-definite matrix $M$ with the sparsity pattern
		\begin{equation}
			M = \begin{bmatrix}
			M_{AA} & M_{AB} & 0\\
			M_{AB}^\top & M_{BB} & M_{BC}\\
			0 & M_{BC}^\top & M_{CC}
			\end{bmatrix},
		\end{equation}
		its unique max-det completion has the following form
		\begin{equation}
			\bar{M} = \begin{bmatrix}
			M_{AA} & M_{AB} & M_{AB}M_{BB}^{-1}M_{BC}\\
			M_{AB}^\top & M_{BB} & M_{BC}\\
			(M_{AB}M_{BB}^{-1}M_{BC})^\top & M_{BC}^\top & M_{CC}
			\end{bmatrix}
		\end{equation}
	\end{lemma}

\vspace{2mm}
\begin{proof}
	The proof can be found in~\cite{Fukuda01} and is omitted for brevity.
\end{proof}

\vspace{2mm}

Without loss of generality, we assume that the perfect elimination ordering of every chordal graph in the sequel is $(1,2,...,d)$.

\vspace{2mm}
\begin{lemma}\label{lemma3}
	Given $M\in\mathbb{S}^d_{++}$, suppose $\text{supp}(M)$ is chordal. For a given $k$ between $1$ and $d-1$, define
	\begin{subequations}
		\begin{align}
			& A_k = \{k\}\\
			& B_k = \{k+1,...,d\}\cap\mathcal{N}(k)\\
			& C_k = \{1,...,d\}\backslash (A\cup B)
		\end{align}
	\end{subequations}
	Then, one can write
	\begin{equation}\label{recursive}
		\bar{M}_{A_kC_k} = M_{A_kB_k}(\bar{M}^{k+1}_{B_kB_k})^{-1}\bar{M}_{B_kC_k}^{k+1}
	\end{equation}
\end{lemma}

\vspace{2mm}
\begin{proof}
	The proof is the immediate consequence of Lemma~\ref{lemma2} and Lemma 2.7 in~\cite{Fukuda01}. The details are omitted for brevity.
\end{proof}
\vspace{2mm}

Lemma~\ref{lemma3} introduces a recursive method for finding the max-det completion of a matrix with chordal structure: given the the $k+1^{\text{th}}$ order max-det completion of $M$, one can find the missing elements in the $k^{\text{th}}$ order max-det completion of $M$ via~\eqref{recursive}. Therefore, $\bar{M}$ can be obtained after performing $d$ iterations of~\eqref{recursive} in the reverse perfect elimination order. 

\vspace{2mm}
\begin{lemma}\label{lemma5}
	Given a positive-definite matrix $M$ and its max-det completion $\bar{M}$, we have $M^{(c)} = \bar{M}-M$.
\end{lemma}
\begin{proof}
	The proof is the direct consequence of Lemma 8 in~\cite{Salar17}. The details are omitted for brevity.
\end{proof}

\vspace{2mm}
Lemma~\ref{lemma5} shows that the nonzero elements of the inverse consistent complement of $M$ are equal to the missing elements in its max-det completion. Therefore, in order to derive an upper bound on $\beta(\mathcal{G},\alpha)$, it suffices to analyze the behavior of the max-det completion of every $M\in\mathbb{S}^d_{++}$ with the diagonal elements equal to 1 that satisfies $\text{supp}(M) = \mathcal{G}$ and $\|M\|_{\max}\leq\alpha$.

\vspace{2mm}
\textit{Proof of Theorem~\ref{thm3}:} Due to Lemmas~\ref{lemma3} and~\ref{lemma5}, it suffices to show that $\|\bar{M}_{A_kC_k}\|_{\max}$ is upper bounded by the right-hand side of~\eqref{upper} for every $k\in\{1,2,...,d\}$ and every $M\in\mathbb{S}^d_{++}$ such that $\text{supp}(M) = \mathcal{G}$ and $\|M\|_{\max}\leq\alpha$. We show this by induction on the order of the max-det completion of $M$. The base case can be easily verified. Suppose that aforementioned upper bound holds for $\|\bar{M}_{A_{j}C_{j}}\|_{\max}$, where $j\in\{k+1,...,d\}$. This implies that 
\begin{equation}\label{eq24}
	\|\bar{M}^{k+1}\|_{\max}\leq \frac{{w}(\mathcal{G})\sqrt{d-{w}(\mathcal{G})-1}\times\alpha^2}{1-(w(\mathcal{G})-1)\times\alpha}\leq\alpha
\end{equation} 
where the second inequality is due to Lemma~\ref{lemma22}. One can write
\begin{align}
	\!\!\!\|\bar{M}_{A_kC_k}\|_{\max} &{=} \|M_{A_kB_k}(\bar{M}^{k+1}_{B_kB_k})^{-1}\bar{M}_{B_kC_k}^{k+1}\|_{\max}\nonumber\\
	&\leq\|M_{A_kB_k}(\bar{M}^{k+1}_{B_kB_k})^{-1}\bar{M}_{B_kC_k}^{k+1}\|_{2}\nonumber\\
	&\leq\|M_{A_kB_k}\|_2\|(\bar{M}^{k+1}_{B_kB_k})^{-1}\|_2\|\bar{M}_{B_kC_k}^{k+1}\|_2\nonumber\\
	&
	\leq\frac{|B_k|\sqrt{|A_k| |C_k|}\times\alpha^2}{\lambda_{\min}(\bar{M}^{k+1}_{B_kB_k})}
\end{align}
where the first inequality is due to Lemma~\ref{lemma3} and the rest of the inequalities are based on the properties of $\max$- and 2-norms. Now, note that $\|\bar{M}^{k+1}_{B_kB_k})\|_{\max}$ is upper bounded by $\alpha$ due to~\eqref{eq24}, $\alpha<1/(w(\mathcal{G})-1)$ and $|B_k|\leq w(\mathcal{G})$. Therefore, one can show that $\bar{M}^{k+1}_{B_kB_k}$ is diagonally dominant and Gershgorin circle theorem can be invoked to obtain $\lambda_{\min}(\bar{M}^{k+1}_{B_kB_k})\geq 1-(w(\mathcal{G})-1)\times\alpha$. Furthermore, we have $|A_k| = 1$, $|B_k|\leq w(\mathcal{G})$, and $|C_k| = d-1-|B_k|$. Under Condition 2-ii, one can show that $|B_k|\sqrt{|A_k| |C_k|}\leq {w}(\mathcal{G})\sqrt{d-{w}(\mathcal{G})-1}$. This completes the proof.~\hfill$\blacksquare$

\vspace{2mm}
\textit{Proof of Corollary~\ref{cor1}:} One can easily verify that the right-hand side of Condition 2-iii in Theorem~\ref{thm3} is in the order of $\mathcal{O}(d^{-\frac{1}{2}})$. On the other hand, since $\epsilon-\delta>1/2$ (due to~\eqref{upperep}), there exists $d'$ such that for $\alpha = \sigma_1-\lambda$ and every $d\geq d_0$, Condition 2-iii is satisfied. Furthermore, it can be easily verified that the right-hand side of~\eqref{upper} is in the order of $\mathcal{O}(d^{-(2\epsilon-2\delta-1/2)})$. Combined with~\eqref{sigma1} and~\eqref{upperep}, one can show that there exists $d''$ such that for every $d\geq d''$, we have
\begin{equation}
\frac{{w}(\mathcal{G})\sqrt{d-{w}(\mathcal{G})-1}\times(\sigma_1-\lambda)^2}{1-(w(\mathcal{G})-1)\times(\sigma_1-\lambda)}\leq \lambda-\sigma_{k+1}
\end{equation}
The proof is completed by setting $d_0 = \max\{d',d''\}$.~\hfill$\blacksquare$
	
\end{document}